\documentclass[conference,a4paper,10pt]{IEEEtran}
\IEEEoverridecommandlockouts
\usepackage{cite}
\usepackage{amsmath,amssymb,amsfonts}
\usepackage{graphicx}
\usepackage{textcomp}
\usepackage{xcolor}
\def\BibTeX{{\rm B\kern-.05em{\sc i\kern-.025em b}\kern-.08em
    T\kern-.1667em\lower.7ex\hbox{E}\kern-.125emX}}

\usepackage{epsfig}
\usepackage{graphicx}
\usepackage{amssymb,multirow,color}
 \usepackage{threeparttable}

\usepackage{amsfonts,amsmath,enumitem,algorithm,algcompatible,booktabs}      %

\algnewcommand\INPUT{\item[\textbf{Input:}]}%
\algnewcommand\OUTPUT{\item[\textbf{Output:}]}%

\newcommand{\bd}{\boldsymbol}
\newcommand{\mb}{\mathbf}
\newcommand{\be}{\begin{equation}}
	\newcommand{\ee}{\end{equation}}

\DeclareMathOperator*{\argmin}{arg\,min}

 \usepackage{url}
\usepackage{tikz}
\newcommand\copyrighttext{%
	\footnotesize \copyright 2022 IEEE. Personal use of this material is permitted. Permission from IEEE must be
	obtained for all other uses, in any current or future media, including
	reprinting/republishing this material for advertising or promotional purposes, creating new
	collective works, for resale or redistribution to servers or lists, or reuse of any copyrighted
	component of this work in other works.}
\newcommand\copyrightnotice{%
	\begin{tikzpicture}[remember picture,overlay]
		\node[anchor=south,yshift=10pt] at (current page.south) {\fbox{\parbox{\dimexpr\textwidth-\fboxsep-\fboxrule\relax}{\copyrighttext}}};
	\end{tikzpicture}%
}
\begin{document}

\title{mFI-PSO: A Flexible and Effective Method\\
	 in Adversarial Image Generation for \\ Deep Neural Networks
 \thanks{Dr. Ziqi Chen's work is partially supported by National Natural Science Foundation of China (Grant No. 11871477) and Natural Science Foundation of Shanghai (Grant No. 21ZR1418800). Dr. Hai Shu's work is partially supported by a startup fund from New York University.}
 }

\author{\IEEEauthorblockN{Hai Shu\textsuperscript{*}, Ronghua Shi\textsuperscript{\dag}, Qiran Jia\textsuperscript{*}, Hongtu Zhu\textsuperscript{\ddag}, Ziqi Chen\textsuperscript{\S}}
\IEEEauthorblockA{\textsuperscript{*}Department of Biostatistics, School of Global Public Health, New York University, New York, NY, USA \\
\textsuperscript{\dag}School of Mathematics and Statistics, Central South University, Changsha, China\\
\textsuperscript{\ddag}Departments of Biostatistics, Statistics, Computer Science, and Genetics,\\
The University of North Carolina at Chapel Hill,
Chapel Hill, NC, USA\\
\textsuperscript{\S}Key Laboratory of Advanced Theory and Application in Statistics and Data Science-MOE,\\
School of Statistics, East China Normal University, Shanghai, China\\
Email: zqchen@fem.ecnu.edu.cn}
}

\maketitle

	\copyrightnotice

\begin{abstract}
Deep neural networks (DNNs) have achieved great success in image classification, but can be very vulnerable to adversarial attacks with small perturbations to images. 
To improve adversarial image generation 
for DNNs, we develop a novel method, called mFI-PSO, which utilizes
a \underline{M}anifold-based \underline{F}irst-order \underline{I}nfluence measure for vulnerable image and pixel selection
and the \underline{P}article \underline{S}warm \underline{O}ptimization for various objective functions. Our mFI-PSO can thus
effectively design adversarial images with flexible, customized options on the number of perturbed pixels, the misclassification probability, and the targeted incorrect class.
Experiments demonstrate the flexibility and effectiveness of our 
mFI-PSO in adversarial attacks 
and its appealing advantages over some popular methods.

\end{abstract}

\begin{IEEEkeywords}
adversarial attack, influence measure, particle swarm optimization, perturbation manifold.
\end{IEEEkeywords}

\section{Introduction}

Deep neural networks (DNNs) have exhibited exceptional performance in image classification \cite{krizhevsky2012imagenet,he2016deep,huang2017densely} and thus are widely used in 
various real-world applications including face recognition \cite{sun2015deepid3}, self-driving cars \cite{bojarski2016end}, biomedical image processing \cite{Lyu21}, among many others \cite{najafabadi2015deep}.
Despite these successes, DNN classifiers can be easily attacked by
adversarial examples with perturbations imperceptible to human vision \cite{szegedy2013intriguing, goodfellow2014explaining, su2019one}.
This motivates the hot research in adversarial attacks and defenses of DNNs \cite{wiyatno2019adversarial,ren2020adversarial}.

Existing adversarial attacks can be categorized into white-box, gray-box, and black-box attacks. Adversaries in white-box attacks have the full information of their targeted DNN model, whereas their knowledge is limited to  model structure  in gray-box attacks and only to model's input and output in black-box attacks. For instance,   
popular algorithms for white-box attacks include the fast gradient sign method (FGSM) \cite{goodfellow2014explaining}, 
the projected gradient descent (PGD) method \cite{madry2017towards},
the Carlini and Wagner (CW) attack \cite{carlini2017towards},
the Jacobian-based saliency map attack (JSMA) \cite{papernot2016limitations}, DeepFool \cite{moosavi2016deepfool},
among many others~\cite{szegedy2013intriguing,kurakin2016adversarial}.

\begin{figure*}[t!]
	\centering
	\includegraphics[width=1\textwidth]{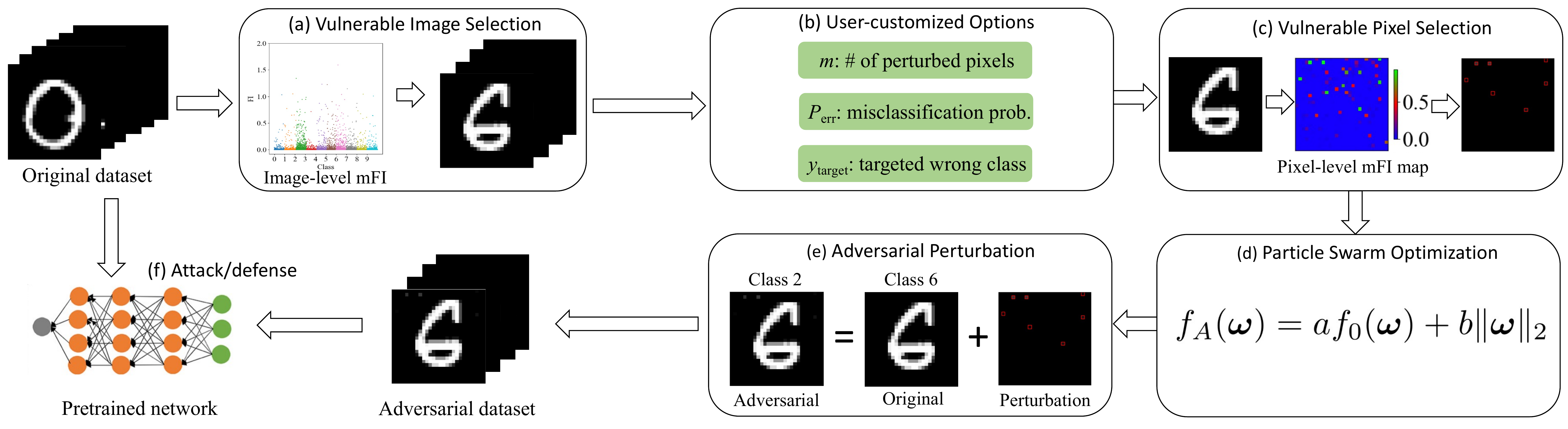}
	\caption{Flowchart of our mFI-PSO method.}
	\label{fig: framework}
\end{figure*}

In this paper, we propose a simple yet efficient method for white-box adversarial image generation 
for DNN classifiers.
For generating an adversarial example of a given image,
our method provides user-customized options on the number of perturbed pixels, misclassification probability, and targeted incorrect class.
To the best of our knowledge, this is the first approach rendering all the three desirable options.

The freedom to specify the number of perturbed pixels
allows users to conduct attacks at various pixel levels
such as one-pixel \cite{su2019one},
few-pixel \cite{papernot2016limitations},
and all-pixel \cite{moosavi2017universal} attacks. 
In particular, we adopt a recent manifold-based first-order influence (mFI) measure 
\cite{shu2019sensitivity} to efficiently locate the most
vulnerable pixels to increase the attack success rate.
Besides, to generate a high-quality adversarial image set, 
our method also utilizes
the mFI measure to identify vulnerable images in a given dataset.
In contrast with traditional Euclidean-space based measures such as Jacobian norm \cite{novak2018sensitivity} and Cook's local influence measure \cite{cook1986assessment}, 
the mFI measure captures the ``intrinsic" change of
the perturbed objective function~\cite{zhu2007perturbation,zhu2011bayesian} and shows better performance in detecting vulnerable images and pixels.

Our method allows users to specify the misclassification probability for
a targeted or nontargeted attack. 
The prespecified misclassification probability is rarely
seen in existing approaches, which produce an adversarial example
either near the model's decision boundary \cite{moosavi2016deepfool,nazemi2019potential} or with unclear confidence \cite{nguyen2015deep}. 
For instance, the CW attack \cite{carlini2017towards} uses a $\kappa$ parameter to control the confidence of an adversarial example, but the parameter is not the misclassification probability that is more user-friendly. In addition, the CW method does not provide the option to control the number of perturbed pixels.

Moreover, we tailor different loss functions accordingly to 
the above three desirable options and their combinations, and apply the particle swarm optimization (PSO) \cite{kennedy1995particle} to obtain the optimal perturbation.
The PSO is a widely-used gradient-free method that is preferred over gradient-based methods
in solving nonconvex or nondifferentiable problems \cite{gudise2003comparison,warsito2019particle},
and thus
provides us much freedom to design various loss functions
to meet users' needs.


Our proposed method is named as mFI-PSO,
based on its combination nature of the mFI measure and the PSO algorithm. 
Figure~\ref{fig: framework} illustrates the flowchart of our method.

We notice that two recent papers \cite{zhang2019attacking,mosli2019they} also applied PSO to craft adversarial images. However, we have intrinsic distinctions. First, the two papers focus on black-box attacks, but ours is white-box.
Article \cite{zhang2019attacking} only studied all-pixel attacks;
although article \cite{mosli2019they} considered few-pixel attacks but searched in random chunks to locate the vulnerable pixels, we use the mFI measure to directly discover those pixels. Moreover, targeted attacks are not considered in \cite{mosli2019they}, and both papers cannot prespecify a misclassification probability for the generated adversarial example.
Our mFI-PSO method is able to design arbitrary-pixel-level, confidence-specified, and/or 
targeted/nontargeted attacks.

Our contributions are summarized as follows:

\begin{itemize}[label=$\bullet$]
	\item We propose a novel white-box method for adversarial image generation 
	for DNN classifiers. It provides users with multiple options 
	on pixel levels, confidence levels, and targeted classes for adversarial attacks. 
	
	\item We innovatively adopt a mFI measure 
	based on an ``intrinsic" perturbation manifold
	to efficiently identify vulnerable images and pixels for adversarial perturbations.
	
	\item We design various different loss functions adaptive to user-customized specifications and apply the PSO, a gradient-free optimization, to obtain optimal perturbations. 
	
	\item We demonstrate the flexibility and effectiveness of our mFI-PSO method in adversarial attacks 
	via experiments on  benchmark datasets and show its winning advantages over
	some commonly-used methods.	
\end{itemize}

The Python code to implement the proposed mFI-PSO is available at \url{https://github.com/BruceResearch/mFI-PSO}.

\section{Method}

\subsection{Manifold-based Influence Measure}

Given an input image $\bd{x}$ and a DNN classifier $N$ with parameters $\bd{\theta}$, the prediction probability for class $y\in \{1,\dots, K\}$ is denoted by $P(y|\bd{x},\bd{\theta})=N_{\bd{\theta}}(y, \bd{x})$.
Let $\bd{\omega}=(\omega_1,\dots,\omega_m)^T$ be a perturbation vector 
in an open set $\Omega\subseteq \mathbb{R}^m$, which can be imposed on any subvector of $(\bd{x}^T,\bd{\theta}^T)^T$. Let the prediction probability under perturbation $\bd{\omega}$ be 
$P(y|\bd{x},\bd{\theta},\bd{\omega})$ with $P(y|\bd{x},\bd{\theta},\bd{\omega}_0)=P(y|\bd{x},\bd{\theta})$
and $\bd{\omega}_0\in \Omega$.

For sensitivity analysis of DNNs, article \cite{shu2019sensitivity} recently has proposed a mFI measure to delineate the {\it intrinsic} perturbed change of the objective function on a Riemannian manifold.
In contrast with traditional Euclidean-space based measures such as Jacobian norm~\cite{novak2018sensitivity} and Cook's local influence measure \cite{cook1986assessment}, 
this perturbation-manifold based measure enjoys 
the desirable {\it invariance} property under diffeomorphic (e.g., scaling) reparameterizations of perturbations 
and has better numerical performance in detecting vulnerable images and pixels.

Let $f(\bd{\omega})$ be an objective function of interest, for example, the cross-entropy 
$f(\bd{\omega})=-\log P(y=y_{\text{true}}|\bd{x},\bd{\theta},\bd{\omega})$.
The mFI measure at $\bd{\omega}=\bd{\omega}_0$ is defined by
\be\label{FI measure}
\text{mFI}_{\bd{\omega}}(\bd{\omega}_0)=
[\partial_{\bd{\omega}} f(\bd{\omega}_0)] \mb{G}^\dag_{\bd{\omega}}(\bd{\omega}_0)  [\partial_{\bd{\omega}} f(\bd{\omega}_0)]^T,
\ee
where  $\partial_{\bd{\omega}}=(\partial/\partial{\omega}_1,\dots,\partial/\partial{\omega}_m)$,
$\mb{G}_{\bd{\omega}}(\bd{\omega})=\sum_{y=1}^K \partial_{\bd{\omega}}^T\ell(\bd{\omega}|y,\bd{x},\bd{\theta})
\partial_{\bd{\omega}}\ell(\bd{\omega}|y,\bd{x},\bd{\theta})P(y|\bd{x},\bd{\theta},\bd{\omega})$
with
$\ell(\bd{\omega}|y,\bd{x},\bd{\theta})=\log P(y|\bd{x},\bd{\theta},\bd{\omega})$, and $\mb{G}^\dag_{\bd{\omega}}(\bd{\omega}_0)$ is the pseudoinverse of $\mb{G}_{\bd{\omega}}(\bd{\omega}_0)$.
A larger value of $\text{mFI}_{\bd{\omega}}(\bd{\omega}_0)$ indicates that
the DNN model is more sensitive in $f(\bd{\omega})$ to local perturbation $\bd{\omega}$ around $\bd{\omega}_0$.

In \eqref{FI measure}, we can see that $\text{mFI}_{\bd{\omega}}(\bd{\omega}_0)$
is an extension of the squared Jacobian norm $\|\mb{J}_{\bd{\omega}}(\bd{\omega}_0)\|_F^2=\partial_{\bd{\omega}} f(\bd{\omega}_0)[\partial_{\bd{\omega}} f(\bd{\omega}_0)]^T$
that is corrected with $\mb{G}^\dag_{\bd{\omega}}(\bd{\omega}_0)$.
When $\mb{G}_{\bd{\omega}}(\bd{\omega}_0)=\mb{I}$, $\text{mFI}_{\bd{\omega}}(\bd{\omega}_0)$ reduces to $\|\mb{J}_{\bd{\omega}}(\bd{\omega}_0)\|_F^2$.
Note that $\mb{G}_{\bd{\omega}}(\bd{\omega})$ is the metric tensor of 
the pseudo-Riemannian manifold $\mathcal{M}=\{P(y|\bd{x},\bd{\theta},\bd{\omega}):\bd{\omega}\in \Omega\}$.
If $\mb{G}_{\bd{\omega}}(\bd{\omega})$ is positive definite, 
then it can define an inner product on $T_{\bd{\omega}}=span(\{\partial \ell(\bd{\omega}|y,\bd{x},\bd{\theta})/\partial \omega_i\}_{i=1}^m)$
which is the tangent space of $\mathcal{M}$ at point $\bd{\omega}$,
and thus can measure the distance of two points on $\mathcal{M}$.
However, $\mb{G}_{\bd{\omega}}(\bd{\omega})$ is a singular matrix when 
the dimension of $\bd{\omega}$ is larger than the number of classes, i.e.,
$m>K$.
The singularity of $\mb{G}_{\bd{\omega}}(\bd{\omega})$
indicates that $\partial \ell/\partial \omega_1,\dots,\partial \ell/\partial \omega_m$, 
which span $T_{\bd{\omega}}$,
are linearly dependent and thus some components of $\bd{\omega}$ are redundant.
Therefore, $\bd{\omega}$ is transformed to a low-dimensional vector $\bd{\nu}$
by
$\bd{\nu}=\mb{\Lambda}_0^{1/2}\mb{U}_0^T \bd{\omega}$, where
$\mb{U}_0\in \mathbb{R}^{m\times r_0}$, with $r_0=\text{rank}(\mb{G}_{\bd{\omega}}(\bd{\omega}_0))$,
and the diagonal matrix $\mb{\Lambda}_0$
form the compact singular value decomposition $\mb{G}_{\bd{\omega}}(\bd{\omega}_0)=\mb{U}_0\mb{\Lambda}_0\mb{U}_0^T$.
Then, $\mathcal{M}_{\bd{\nu}_0}=\{P(y|\bd{x},\bd{\theta},\bd{\nu}):\bd{\nu}\in B_{\bd{\nu}_0}\}$ is a Riemannian manifold with a positive-definite metric tensor $\mb{G}_{\bd{\nu}}(\bd{\nu})$
in some open ball $B_{\bd{\nu}_0}$ centered at $\bd{\nu}_0=\mb{\Lambda}_0^{1/2}\mb{U}_0^T \bd{\omega}_0$, and moreover, $\mb{G}_{\bd{\nu}}(\bd{\nu}_0)=\mb{I}$.
This indicates that $\partial \ell/\partial \nu_1,\dots, \partial \ell/\partial \nu_{r_0}$ 
are a basis of the tangent space, $T_{\bd{\nu}}$, of $\mathcal{M}_{\bd{\nu}_0}$ at $\bd{\nu}$,
and are an orthonormal basis of $T_{\bd{\nu}}$ at $\bd{\nu}=\bd{\nu}_0$.
We can thus view $\bd{\nu}$ as the {\it  intrinsic} representation of perturbation $\bd{\omega}$.
Furthermore, we have that
$
\text{mFI}_{\bd{\omega}}(\bd{\omega}_0)=[\partial_{\bd{\omega}} f(\bd{\omega}_0)] \mb{G}^\dag_{\bd{\omega}}(\bd{\omega}_0)  [\partial_{\bd{\omega}} f(\bd{\omega}_0)]^T
=[\partial_{\bd{\nu}} f(\bd{\nu}_0)] \mb{G}^{-1}_{\bd{\nu}}(\bd{\nu}_0)  [\partial_{\bd{\nu}} f(\bd{\nu}_0)]^T
=\partial_{\bd{\nu}} f(\bd{\nu}_0)[\partial_{\bd{\nu}} f(\bd{\nu}_0)]^T
=\| \mb{J}_{\bd{\nu}}(\bd{\nu}_0)\|_F^2,
$
i.e., the measure $\text{mFI}_{\bd{\omega}}(\bd{\omega}_0)$ is equal to 
the squared Jacobian norm $\| \mb{J}_{\bd{\nu}}(\bd{\nu}_0)\|_F^2$ 
using the intrinsic perturbation~$\bd{\nu}$.
See \cite{shu2019sensitivity} for the detailed derivation
of the definition of $\text{mFI}_{\bd{\omega}}(\bd{\omega}_0)$ in~\eqref{FI measure}
from the Riemannian manifold $\mathcal{M}_{\bd{\nu}_0}$.

We shall apply the mFI measure to discover vulnerable images or pixels for adversarial perturbations. It is worth mentioning that \cite{shu2019sensitivity}
did not develop an optimization algorithm for adversarial attacks
that incorporates their proposed mFI measure.
We will connect 
the mFI measure to adversarial attacks by our devised optimizations 
that can be solved by the PSO algorithm.

\subsection{Particle Swarm Optimization}

Since introduced by \cite{kennedy1995particle},
the PSO algorithm has been successfully used in solving complex optimization problems in various fields of engineering and science \cite{poli2008analysis,shi2001particle,zhang2015comprehensive}.  
Let $f_A$ be an objective function, which will be specified in Section~\ref{sec: AIG} for adversarial scenarios. 
The PSO algorithm performs searching via a population (called swarm) of candidate solutions (called particles) by iterations to optimize the objective function $f_A$. Specifically, let 
\begin{align}
	\bd{p}_{i,{\text{best}}}^t&=\argmin_{k=1,\dots, t} f_A(\bd{\omega}_i^k),\quad i\in \{1,2,\dots, N_p\},\label{PSO 1}\\
	\bd{g}_{\text{best}}^t&=\argmin_{i=1,\dots, N_p} f_A( \bd{p}_{i,{\text{best}}}^t)
	\label{PSO 1.2},
\end{align}
where $\bd{\omega}_i^k=(\omega_{i1}^k,\dots, \omega_{im}^k)^T$ is the position of particle $i$ in an $m$-dimensional space at iteration $k$,  $N_p$ is the total number of particles, and $t$ is the current iteration.
The position, $\bd{\omega}_i^{t+1}$, of particle $i$ at iteration $(t+1)$ is updated with 
a velocity $\bd{v}_i^{t+1}=(v_{i1}^{t+1},\dots,v_{im}^{t+1})$ by
\be\label{PSO 2}
\begin{aligned}
	\bd{v}_i^{t+1}&=w\bd{v}_i^{t}+c_1r_1(\bd{p}_{i,{\text{best}}}^t-\bd{\omega}_i^{t})
	+c_2r_2(\bd{g}_{\text{best}}^t-\bd{\omega}_i^{t}),\\
	\bd{\omega}_i^{t+1}&=\bd{\omega}_i^{t}+\bd{v}_i^{t+1},
\end{aligned}
\ee
where $w$ is the inertia weight, $c_1$ and $c_2$ are acceleration coefficients, and $r_1$ and $r_2$ are uniformly distributed random variables in the range $[0,1]$.
Following \cite{xu2019particle}, we fix $w=0.6$ and $c_1=c_2=2$.
The movement of each particle is guided by 
its individual best known position and the entire swarm's best known position.

The PSO is a widely-used gradient-free method that is 
more stable and efficient
than gradient-based methods
to solve nonconvex or nondifferentiable problems \cite{gudise2003comparison,warsito2019particle}.
This motivates us to 
adopt the PSO to optimize our various objective functions (in Section \ref{sec: AIG}),
designed to meet different user's requirements on adversarial images,
which may not be convex or differentiable.

\subsection{Adversarial Image Generation}\label{sec: AIG}

Given an image $\bd{x}$, 
we innovatively combine the mFI measure and the PSO to generate its adversarial image with user-customized options on 
the number of pixels for perturbation, the misclassification probability, and  the targeted class to which the image is misclassified, denoted by $m$, 
$P_{\text{err}}$, and $y_{\text{target}}$, respectively.

Denote image $\bd{x}=(x_1,\dots, x_p)^T$. 
For an RGB image of $q$ pixels, we view the three channel components of a pixel as three separate pixels, so $p=3q$ here. We let the default value of $m=p$.

We first locate $m$ vulnerable pixels in $\bd{x}$ for perturbation,
if $m$ is specified but the targeted pixels are not given by the user.
We compute the mFI measure in \eqref{FI measure} for each pixel $i\in\{1,\dots, p\}$ based on the objective function
\be\label{f for each pixel}
f(\bd{\omega})=
\begin{cases}
	-\log P(y_{\text{true}}|\bd{x},\bd{\theta},\bd{\omega}),&\text{if $y_{\text{target}}$ is not given},\\
	-\log P(y_{\text{target}}|\bd{x},\bd{\theta},\bd{\omega}), &\text{otherwise},
\end{cases}
\ee
where $\bd{\omega}=\Delta x_i$.
Denote $x_{(i)}$ to be the pixel with the $i$-th largest mFI value.
We use $x_{(1)},\dots, x_{(m)}$ as the $m$ pixels for adversarial attack
and let
perturbation $\bd{\omega}=(\omega_1,\dots, \omega_m)^T=(\Delta x_{(1)},\dots, \Delta x_{(m)})^T$.

We then apply the PSO in \eqref{PSO 1}--\eqref{PSO 2} to obtain an optimal value of $\bd{\omega}$ that minimizes the adversarial objective function
\[
f_{A}(\bd{\omega})=a f_0(\bd{\omega})+b\|\bd{\omega}\|_2, \quad \omega_i\in \epsilon\cdot [0-x_{(i)},1-x_{(i)}],
\]
where we assume $x_{(i)}\in [0,1]$,
$\epsilon$ constrains the range of perturbation to guarantee the visual quality of the generated adversarial image compared 
to the original, $f_0(\bd{\omega})$ is a misclassification loss function, $\|\bd{\omega}\|_2$ represents the magnitude of perturbation, and $a$ and $b$ are prespecified weights. 
To ensure the misleading nature of the generated adversarial sample, $a\gg b$ is set to
prioritize $f_0(\bd{\omega})$ over  $\|\bd{\omega}\|_2$.

We design different $f_0(\bd{\omega})$ functions to meet different user-customized requirements on $\{m, P_{\text{err}}, y_{\text{target}}\}$.
If only $m$ is known, inspired by \cite{meng2018generate} and \cite{meng2017magnet}, we let the misclassification loss function be
\[
f_0(\bd{\omega})=
\begin{cases}
	\left|P(y_1|\bd{x},\bd{\theta},\bd{\omega})-P(y_2|\bd{x},\bd{\theta},\bd{\omega})\right|,&\text{if $y_1=y_{\text{true}}$},\\
	0, &\text{if $y_1\ne y_{\text{true}}$},
\end{cases}
\]
where $y_k$ is the label with the $k$-th largest prediction probability $P(y=y_k|\bd{x},\bd{\theta},\bd{\omega})$ from the trained DNN 
for the input image $\bd{x}$ added with perturbation $\bd{\omega}$.
Since $y_1\ne y_{\text{true}}$ results in the minimum of $f_0(\bd{\omega})$, 
this loss function encourages PSO to yield a valid perturbation.
If the $\bd{\omega}$-perturbed $\bd{x}$ is prespecified with
a misclassification probability 
$P_{\text{err}}\ge 0.5$, we
use the misclassification loss function
\[
f_0(\bd{\omega})=
\begin{cases}
	\left|P(y_2|\bd{x},\bd{\theta},\bd{\omega})-P_{\text{err}}\right|,\quad &\text{if $y_1=y_{\text{true}}$},\\
	\left|P(y_1|\bd{x},\bd{\theta},\bd{\omega})-P_{\text{err}}\right|, \quad&\text{if $y_1\ne y_{\text{true}}$}.
\end{cases}
\]
If a targeted class $y_{\text{target}}$ is given, we choose the misclassification loss function 
\[
f_0(\bd{\omega})\\
{=}\begin{cases}
	\left|P(y_1|\bd{x},\bd{\theta},\bd{\omega}){-}P(y_{\text{target}}|\bd{x},\bd{\theta},\bd{\omega})\right|,&\text{if $y_1\ne y_{\text{target}}$},\\
	0, &\text{if $y_1= y_{\text{target}}$}.
\end{cases}
\]
Furthermore, if both $P_{\text{err}}$ and $y_{\text{target}}$ are provided, we use 
\[
f_0(\bd{\omega})\\
{=}
\begin{cases}
	\left|P(y_1|\bd{x},\bd{\theta},\bd{\omega}){-}P(y_{\text{target}}|\bd{x},\bd{\theta},\bd{\omega})\right|,&\text{if $y_1\ne y_{\text{target}}$},\\
	\left|P(y_1|\bd{x},\bd{\theta},\bd{\omega})-P_{\text{err}}\right|, &\text{if $y_1= y_{\text{target}}$}.
\end{cases}
\]
or equivalently $f_0(\bd{\omega})=\left|P(y_{\text{target}}|\bd{x},\bd{\theta},\bd{\omega})-P_{\text{err}}\right|$.

Our procedure for generating a customized adversarial image is 
illustrated in Figure~\ref{fig: framework} (b)-(e) and also 
summarized in
Algorithm~\ref{img generation}.

\begin{algorithm}[h!]
	\caption{Adversarial image generation}
	\begin{algorithmic}[1]
		\INPUT Image and label $\{\bd{x},y_{\text{true}}\}$, number of perturbed pixels $m$, (optional) indices of perturbed pixels, (optional) misclassification probability $P_{\text{err}}$, (optional) targeted incorrect label $y_{\text{target}}$, hyperparameters $\{N_p,a,b,\epsilon\}$ in PSO, and maximum iteration number $T$	
		\STATE If perturbed pixels are not specified,
		compute mFI by \eqref{FI measure} and \eqref{f for each pixel} for all pixels to locate the $m$ pixels for perturbation $\bd{\omega}$;
		\STATE Initialize $N_p$ particles in PSO with positions $\{\{ \bd{p}_{i,\text{best}}^0=\bd{\omega}_i^0\}_{i=1}^{N_p},\bd{g}_{\text{best}}^0\}$ and velocities $\{\bd{v}_i^0\}_{i=1}^{N_p}$;
		\STATE {\bf Repeat}
		\STATE \quad {\bf for} particle $i=1,\dots, N_p$ {\bf do}
		\STATE \qquad Update $\bd{v}_i^t$ and $\bd{\omega}_i^t$ by \eqref{PSO 2};
		\STATE \qquad Update $\bd{p}_{i,\text{best}}^t$  by \eqref{PSO 1};
		\STATE \quad {\bf end for}
		\STATE \quad Update $\bd{g}_{\text{best}}^t$ by \eqref{PSO 1.2};
		\STATE 	{\bf Until} convergence or iteration $t=T$
		\OUTPUT Adversarial image $(\bd{x}+\text{zero-padded}~\bd{\omega})$, where $\bd{\omega}= \bd{g}_{\text{best}}^t$.
	\end{algorithmic}
	\label{img generation}
\end{algorithm}


We now aim to create a set of adversarial images for a given trained DNN model.
To include as many adversarial images as possible, one may not need to specify a value to $P_{\text{err}}$ in Algorithm~\ref{img generation}.
Note that Algorithm~\ref{img generation} may not have a feasible solution when given with restrictive parameters such as small $\epsilon$ or small $N_p$.
To efficiently generate a batch of adversarial images, we
first select a set of potentially vulnerable images by
some modifications to Algorithm~\ref{img generation}.

Specifically, given an image dataset $X=\{\bd{x}_i\}_{i=1}^n$, thresholds $\{\text{mFI}_{\text{img}}, P_{\text{target}},$
$\text{mFI}_{\text{pixel}}\}$, 
optional $\{m,P_{\text{err}}\}$, and
targeted incorrect labels $\{y_{i,\text{target}}\}_{i=1}^n$ (if not given, $y_{i,\text{target}}=y_{i,2}$ the label with the second largest prediction probability), we first find $\widetilde{X}$, the set of all correctly classified images that have image-level mFI (with $\bd{\omega}=\Delta\bd{x}_i$) $\ge \text{mFI}_{\text{img}}$ and prediction probability $P(y_{i,\text{target}}|\bd{x}_i,\bd{\theta})\ge P_{\text{target}}$.
For each image in set $\widetilde{X}$, we generate its adversarial image by Algorithm~\ref{img generation} with 
$m$, if not specified,  being the number of pixels with mFI $\ge \text{mFI}_{\text{pixel}}$,
optional misclassification probability $P_{\text{err}}$,
and 
$y_{\text{target}}=y_{i,\text{target}}$.
These generated adversarial images form an adversarial dataset.
The algorithm to generate such an adversarial dataset is detailed in Algorithm~\ref{set generation}. 

\begin{algorithm}[h!]
	\caption{Adversarial dataset generation}
	\begin{algorithmic}[1]
		\INPUT Image set $X=\{\bd{x}_i\}_{i=1}^n$ and labels $\{y_{i,\text{true}}\}_{i=1}^n$, thresholds $\{\text{mFI}_{\text{img}}, P_{\text{target}},$
		$\text{mFI}_{\text{pixel}} \}$, 
		optional $\{m, P_{\text{err}}\}$, 
		targeted incorrect labels $\{y_{i,\text{target}}\}_{i=1}^n$ (if not given, let $y_{i,\text{target}}=y_{i,2}$), and hyperparameters $\{N_p,a,b,\epsilon,T\}$ in Algorithm~\ref{img generation}	
		\STATE For each correctly classified $\bd{x}_i\in X$, compute the image-level mFI (denoted by $\text{mFI}_i$)  by \eqref{FI measure} with
		$\bd{\omega}=\Delta{\bd{x}_i}$ and $f(\bd{\omega})=-\log P(y=y_{i,\text{true}}|\bd{x}_i,\bd{\theta},\bd{\omega})$;		
		\STATE Determine $\widetilde{X}=\{\bd{x}_i\in X: \text{mFI}_i\ge \text{mFI}_{\text{img}},$
		$
		P(y_{i,\text{target}}|\bd{x}_i,\bd{\theta})\ge P_{\text{target}}\}$;
		\STATE For each $\bd{x}_i\in \widetilde{X}$, generate its adversarial image $\bd{x}_i^a$ by Algorithm 1 with 
		$m$, if not given, being \# of pixels with mFI $\ge \text{mFI}_{\text{pixel}}$,
		optional $P_{\text{err}}$, and 
		$y_{\text{target}}=y_{i,\text{target}}$.
		\OUTPUT Adversarial dataset $\widetilde{X}^a=\{\bd{x}_i^a\}_{i=1}^{|\widetilde{X}|}$
	\end{algorithmic}
	\label{set generation}
\end{algorithm}

\section{Experiments}
We conduct experiments on the two benchmark datasets MNIST 
and CIFAR10 
using the ResNet32 model~\cite{he2016deep}. 
Data augmentation is used, including random horizontal and vertical shifts up to 12.5\% of image height and width for both datasets, and additionally random horizontal flip for CIFAR10. 
We train the ResNet32 on MNIST and CIFAR10 for 80 and 200 epochs, respectively. For each dataset, we randomly select 1/5 of training images as the validation set to monitor the training process. Table~\ref{acc table} shows the prediction accuracy of our trained ResNet32 for the two datasets.

We first showcase the proposed mFI-PSO method in Section~\ref{sec: ex mFI-PSO} and then compare it with some widely used methods in Section~\ref{sec: comparison related}.

\begin{table}[h!]
	\caption{Classification accuracy (in \%)
		of originally trained ResNet32.}
	\label{acc table}
	\centering
	\scalebox{1}{
		\begin{tabular}{ccccccccc}
			\toprule
			&	\multicolumn{3}{c}{MNIST/CIFAR10}        \\
			\cmidrule(r){2-4}
			&		Training (n=60k/50k)   & & Test (n=10k/10k)    \\
			\midrule
			&	99.76/98.82  && 99.25/91.28  \\
			\bottomrule
		\end{tabular}
	}
\end{table}

\begin{figure*}[h!]	
	\centering
	\includegraphics[width=0.7\textwidth]{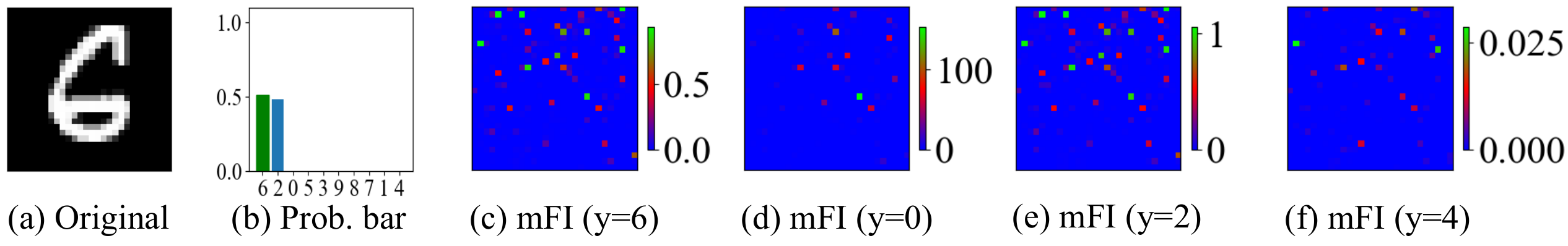}
	\caption{Pixel-level mFI maps of an MNIST image for different target classes.}
	\label{FI MNIST}	
\end{figure*}

\begin{figure*}[h!]
	\centering
	\includegraphics[width=0.65\textwidth]{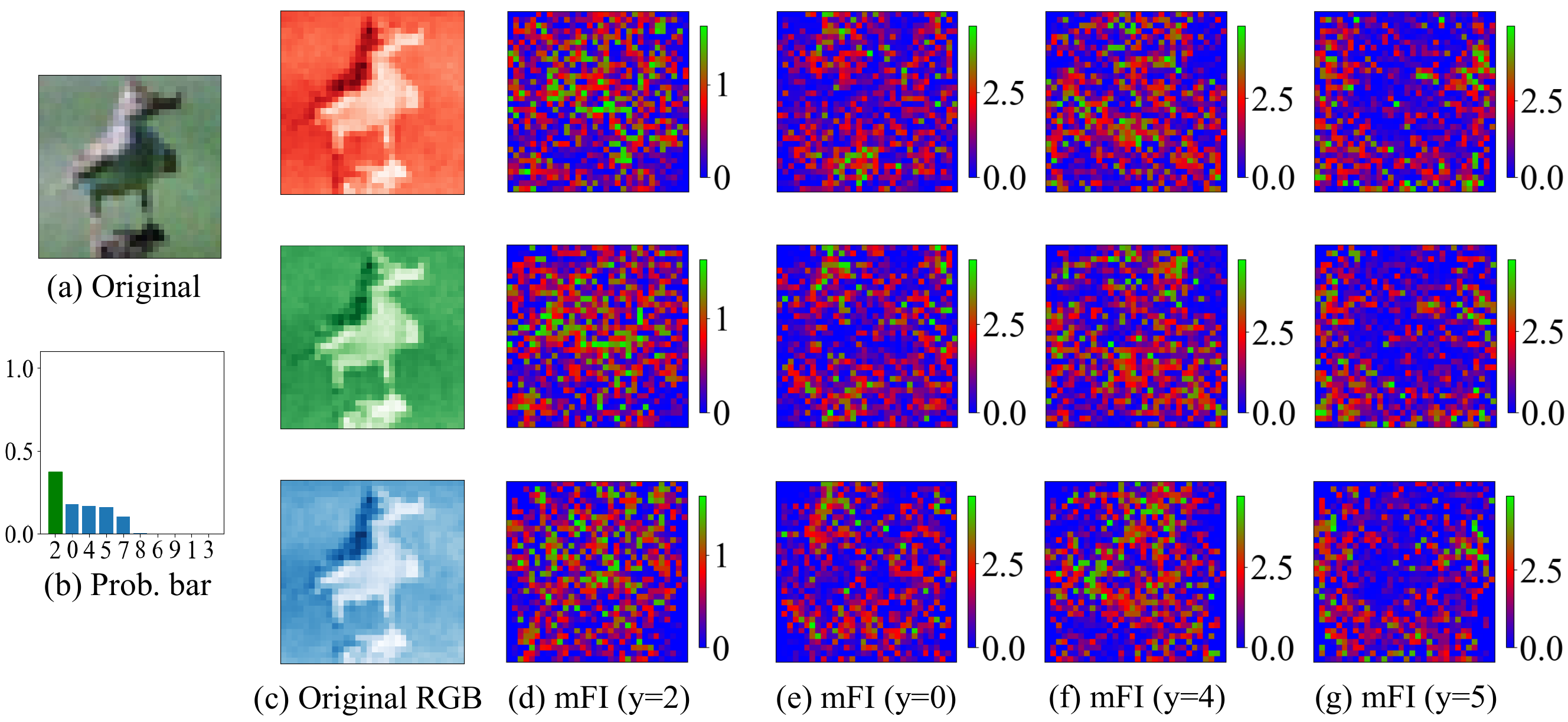}
	\caption{Pixel-level mFI maps of a CIFAR10 image's RGB channels for different target classes. Class labels: (0, 1, 2, 3, 4, 5, 6, 7, 8, 9) = (plane, car, bird, cat, deer, dog, frog, horse, ship, truck).}
	\label{FI CIFAR}
\end{figure*}

\subsection{Illustration of the proposed mFI-PSO}\label{sec: ex mFI-PSO}

We first consider two images with easy visual detection and
large image-level mFI
in MNIST and CIFAR10, shown in Figures~\ref{FI MNIST}-\ref{FI CIFAR} with predictive probability graphs and pixel-level mFI maps. 
The probability bar graphs imply candidate misclassification classes that can be used as $y_{\text{target}}$.
The mFI maps indicate the vulnerability of each pixel to local perturbation and are useful to locate pixels for attack.

\begin{figure*}[h!]
	\centering
	\includegraphics[width=0.58\textwidth]{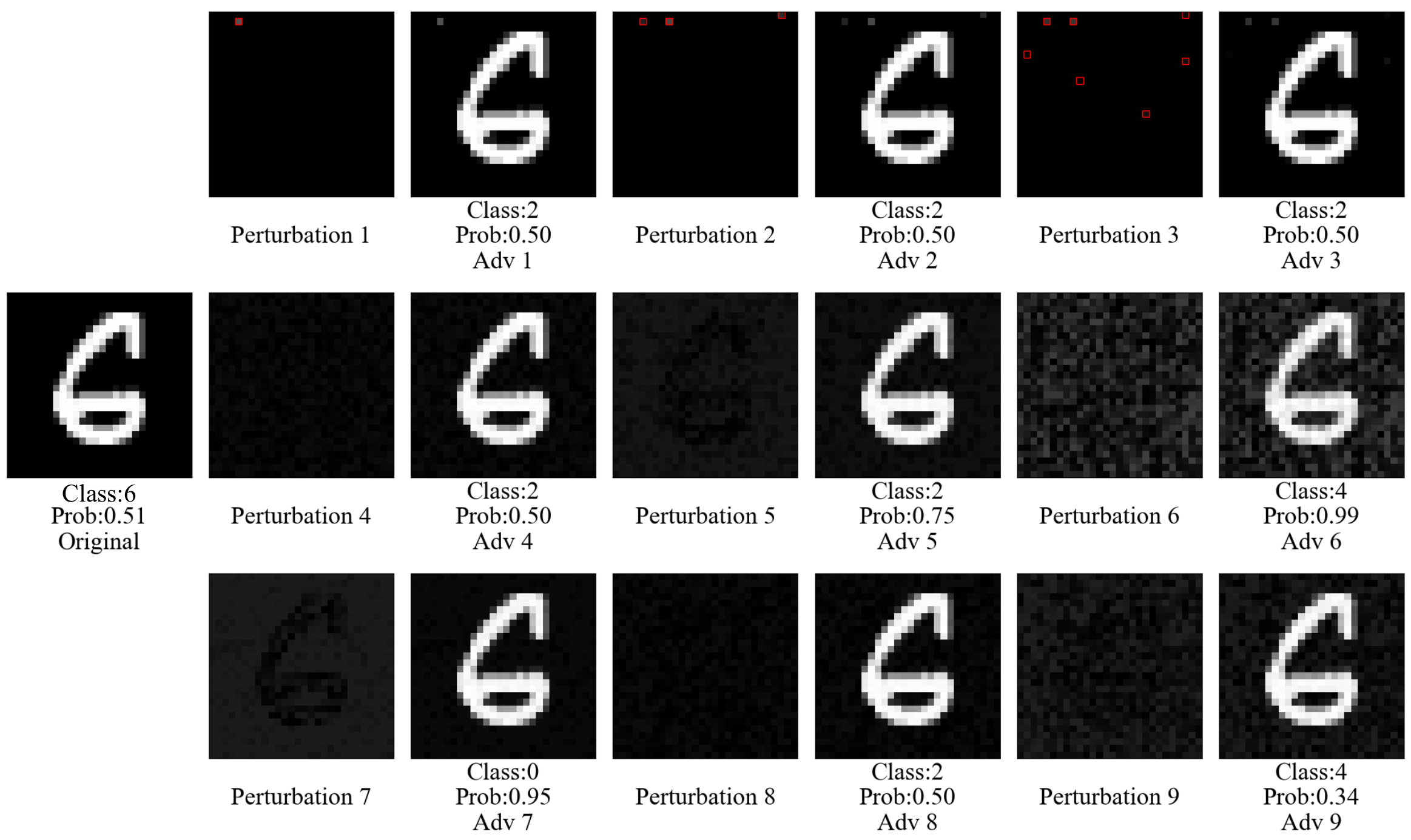}

	\caption{Adversarial examples of an MNIST image. Perturbations\,1-3 are set with $m=1,3,7$, respectively; Perturbations\,4-6 are with $P_{\text{err}}=0.5,0.75,0.99$, respectively; Perturbations\,7-9 are with $y_{\text{target}}=0,2,4$, respectively. Perturbation maps are followed by adversarial images.}
	\label{AD MNIST}
\end{figure*}

\begin{figure*}[h!]
	\centering
	\includegraphics[width=0.58\textwidth]{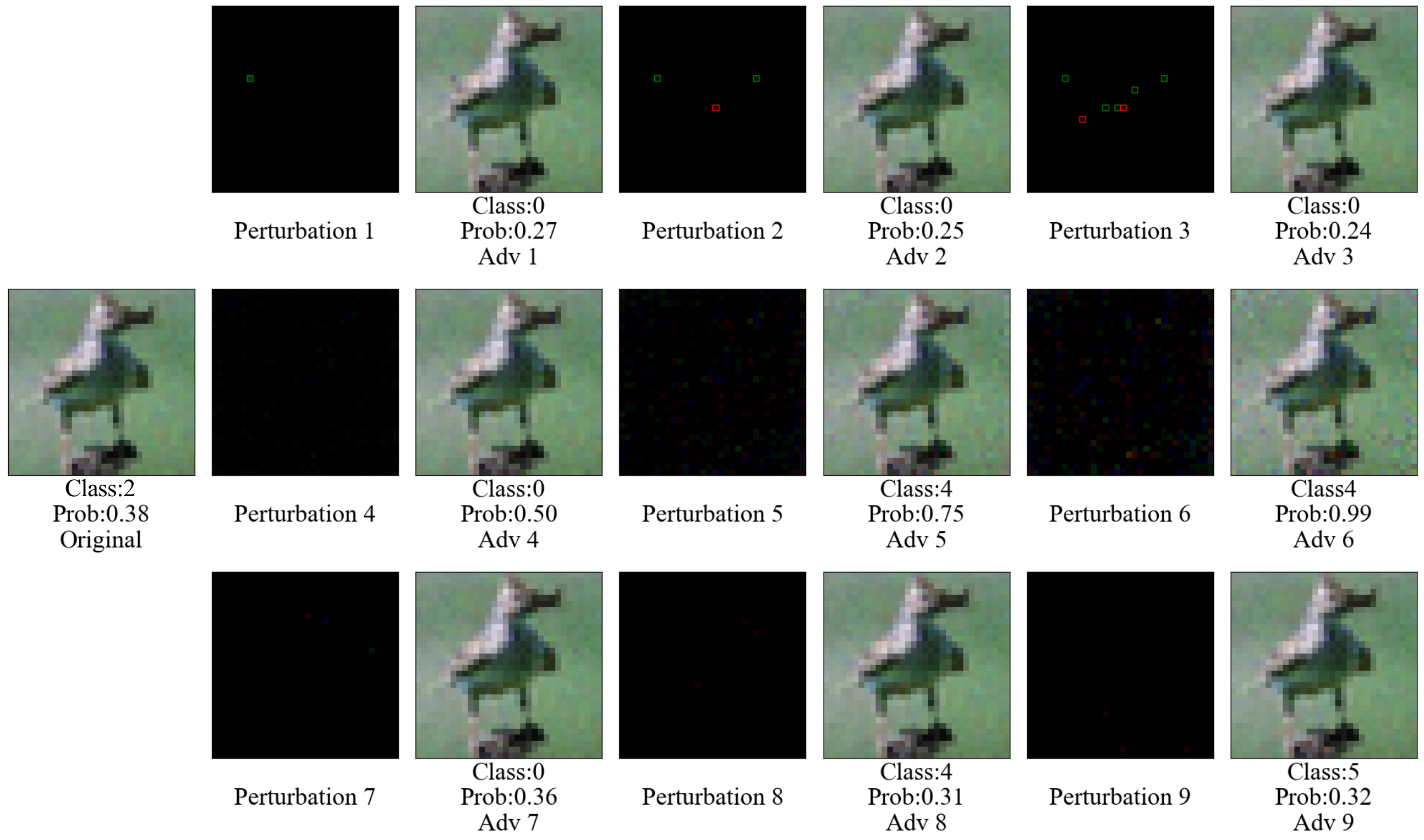}
	
	\caption{Adversarial examples of a CIFAR10 image. Perturbations\,1-3 are set with $m{=}1,3,7$ attacked pixels (framed in the attacked channel's color), respectively. Perturbations\,4-6 are set with $P_{\text{err}}{=}0.5,0.75, 0.99$, respectively. Perturbations\,7-9 are set with $y_{\text{target}}{=}0,4,5$, respectively. Perturbation maps are followed by 
		adversarial images.}
	\label{AD CIFAR}
\end{figure*}

We evaluate the performance of Algorithm\,\ref{img generation} (see Figure\,\ref{fig: framework}\,(b)-(e))
in generating adversarial examples of the two images 
according to different requirements on $m$, $P_{\text{err}}$ and $y_{\text{target}}$.
Figures~\ref{AD MNIST}-\ref{AD CIFAR} show the
generated adversarial images with corresponding perturbation maps. 
Perturbations~1-3 consider the settings with $m=1, 3, 7$, respectively, and with
no specifications to $P_{\text{err}}$ and $y_{\text{target}}$.
For Perturbations~4-6, 
we only specify $P_{\text{err}}=0.5, 0.75, 0.99$, respectively, 
assign no value to $y_{\text{target}}$, and 
tune $m$ being the number of pixels with mFI $\ge\text{mFI}_{\text{pixel}}\in \{0.1,1\}$ and $N_p\in \{200,500,1000\}$ to obtain feasible solutions from PSO.
Perturbations~7-9 are prespecified with $y_{\text{target}}=0, 2, 4$ for MNIST, and $0, 4, 5$ for CIFAR10, respectively, $m$ being the number of pixels with mFI $\ge 0.1$,
and no value for $P_{\text{err}}$.
The generated adversarial images have visually negligible  
differences from the originals and satisfy the prespecified requirements.

\begin{figure*}[h!]
	\centering
	\includegraphics[width=0.75\textwidth,height=3.5cm]{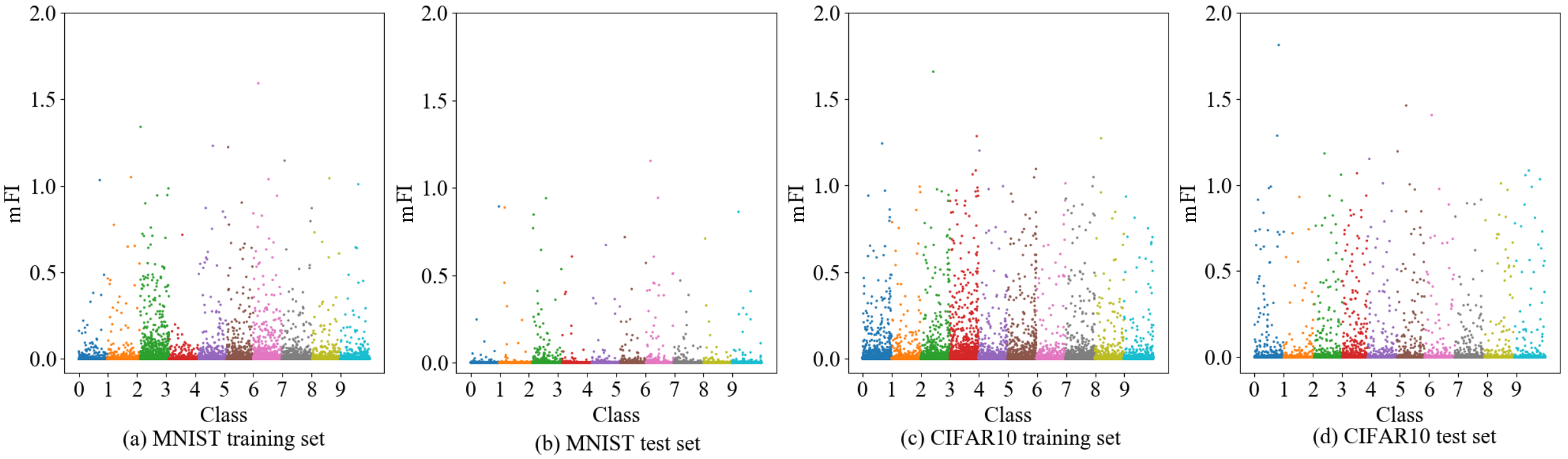}
	\caption{Manhattan plots of image-level mFIs for correctly classified images.}
	\label{Manplot}
\end{figure*}

We now consider using Algorithm~\ref{set generation} to generate adversarial datasets.
Figure~\ref{Manplot} shows the Manhattan plots of image-level mFIs for correctly classified images.
Based on the figure, 
for both MNIST and CIFAR10,
we generate three adversarial datasets by
selecting vulnerable images with image-level
$\text{mFI}\ge\text{mFI}_{\text{img}}=0.2, 0.1, 0.01$, respectively (see Figure~\ref{fig: framework}(a)).
For all the sets,
we set 
$P_{\text{target}}=0.2$ and $\text{mFI}_{\text{pixel}}=0.01$
in Algorithm~\ref{set generation}; 
following \cite{dabouei2020exploiting} we set the bound of the max absolute value (i.e., the $\ell_\infty$ norm) of perturbations $\epsilon=0.15$ for pixel values converted onto $[0,1]$.
Table~\ref{attack succ table} shows
that the success rate of our mFI-PSO attack
is enhanced as the threshold $\text{mFI}_{\text{img}}$ increases, and it is 100\% 
when $\text{mFI}_{\text{img}}=0.2$ and is 
above 96.5\% for MNIST and 92.5\% for CIFAR10 even when $\text{mFI}_{\text{img}}$ is small as 0.01.

\subsection{Comparison with Related Methods}\label{sec: comparison related}

We compare the proposed mFI-PSO with five commonly-used methods, including FGSM \cite{goodfellow2014explaining}, PGD \cite{madry2017towards}, JSMA \cite{papernot2016limitations}, $\ell_\infty$-norm CW (CW$_\infty$) \cite{carlini2017towards}, and DeepFool \cite{moosavi2016deepfool}.
A brief summary of the five existing methods can be found in \cite{ren2020adversarial}.
None of the five methods 
can include all of the aforementioned three desirable user-customized options for crafting adversarial images. 
Moreover, none of them consider the perturbation intrinsically from the Riemannian manifold. 
In particular, similar to our pixel-level mFI map,
a saliency map but based on the Jacobian matrix is utilized in the JSMA. However, as shown in \cite{shu2019sensitivity}, the mFI measure outperforms the Jacobian information in detecting vulnerable images and pixels.

We implement the five previous methods in Python using the {\it Adversarial Robustness Toolbox (ART)} from 
\mbox{\url{https://github.com/Trusted-AI/adversarial-robustness-toolbox}}.
Letting pixel values transform onto $[0,1]$, following \cite{dabouei2020exploiting} we set the bound of the $\ell_\infty$ norm of perturbations $\epsilon=0.15$ for FGSM, PGD, CW$_\infty$, and our mFI-PSO, but JSMA and DeepFool do not have a similar parameter to bound the magnitude of perturbations.  
Our mFI-PSO (Algorithm~2) is set with the same parameters as in Section~\ref{sec: ex mFI-PSO} and 
the other methods are based on the default settings of the {\it ART} package. All methods
are not specified with targeted classes.

\begin{table*}[h!]
	\caption{Attack success rates (in \%) on
		correctly classified images with different image-level mFI thresholds ($\text{mFI}_{\text{img}}$)
		\\
		for originally trained ResNet32. }\vspace{-0.1cm}
	\label{attack succ table}
	\centering
	\begin{threeparttable}
		\begin{tabular}{cc|c|ccc|c|ccccccccccccccccccccccc ccccc cccccc}
			\toprule
			&	\multicolumn{3}{c}{MNIST} &&      \multicolumn{3}{c}{CIFAR10}          \\
			\cline{2-4}\cline{6-8}
			mFI$_{\text{img}}$ value&	0.2& 0.1&0.01&&0.2& 0.1&0.01\\
			CC'ed Tr./Ts.  &	(n=189/46) & (n=413/84) & (n=1876/325) && (n=356/229)& (n=693/385) & (n=2562/1008)\\
			\midrule
			mFI-PSO & {\bf 100/100} & {\bf 100/100} & 96.54/97.54 && {\bf 100/100} &97.26/94.81 &{\bf 92.55/92.66}\\
			FGSM	& 77.78/89.13 & 82.08/91.67 & 81.82/90.46 && {\bf 100/100} & 97.40/98.96 &53.16/64.19\\
			PGD &  99.47/{\bf 100} & 99.76/{\bf 100} & {\bf 98.61/98.38} && {\bf 100/100} & {\bf 97.98/99.74} &58.70/68.35\\
			CW$_\infty$  & 10.05/{\bf 100} & 10.9/{\bf 100} & 9.22/97.54 && 79.21/76.86 & 65.95/63.64 &33.33/42.16\\
			\multirow{2}{*}{JSMA} & 0/0 & 0/0 & 0.053/1.53 && 63.20/60.34 & 65.95/57.92 &57.92/64.48\\
			& (100)/(100) & (100)/(100) & (100)/(100) && (100)/(100) & (100)/(100) & (100)/(100)\\
			\multirow{2}{*}{DeepFool} & 0.53/54.34 & 1.45/69.04 & 1.33/32.62 && 60.34/44.96 & 23.09/27.79 &6.25/10.62\\
			& (92.06)/(56.52) & (94.19)/(70.24) & (95.20)/(90.46) && (100)/(100) & (100)/(100) & (100)/(100)\\
			\bottomrule
		\end{tabular}

		\begin{tablenotes}[flushleft]
			\item 		Note: The preset bound of the $\ell_\infty$ norm of perturbations is
			$\epsilon=0.15$ for pixel values converted onto $[0,1]$. 	
			JSMA and DeepFool do not have the parameter $\epsilon$, and their success rates without 
			the $\ell_\infty$-norm bound are given in the parentheses. 
			Tr. =  training data, Ts. =  test data, CC'ed = correctly classified.
		\end{tablenotes}
	\end{threeparttable}
\end{table*}

\begin{table*}[h!]
	\caption{\centering Classification results of adversarially fine-tuned ResNet32. }
	\label{0.01acc adv table}
	\centering
	\begin{threeparttable}
		\begin{tabular}{r|c|c|cccccccccccccccccccc}
			\toprule
			\multicolumn{4}{c}{MNIST/CIFAR10}           \\
			\hline
			Fine-tuned by Tr. 	 &		Accuracy in \% on     & Accuracy in \% on   & 
			n of CC'ed Ts.
			\\
			(n=60k/50k)+Adv. Tr. of&  Ts. (n=10k/10k)   &  Adv. Ts.  &  with mFI$\ge$0.01 
			
			\\
			\midrule
			mFI-PSO (n=1811/2371)	& {\bf 99.78}/91.63 & {\bf 93.06}(n=317)/65.42(n=934) & {\bf 132}/{\bf 441} \\
			FGSM (n=1535/1362)&{99.65}/{\bf 91.77} & 84.69(n=294)/71.72(n=647)  & 169/570 \\
			PGD (n=1850/1504)& 99.66/91.43 & 82.35(n=323)/71.55(n=689)&  163/603\\
			CW$_\infty$ (n=173/777)& {99.72}/{91.54} & 93.38(n=317)/{72.24}(n=425) & 199/615\\
			JSMA (n=1876/2562)&99.68/91.68 & 91.38(n=325)/{\bf 75.79}(n=1008)&  153/551\\
			DeepFool (n=1786/2556)& 99.60/{\bf 91.77} & 85.03(n=294)/24.80(n=1002) & 368/800 \\
			\bottomrule
		\end{tabular}
		\begin{tablenotes}[flushleft]
			\item Note:  The adversarial training and test sets for each method are selected with image-level mFI $\ge 0.01$. 
			Adv. = adversarial, Tr. =  training data, Ts. =  test data, CC'ed = correctly classified.
		\end{tablenotes}
	\end{threeparttable}
\end{table*}

\begin{table*}[h!]
	\caption{ Attack success rate (in \%) on adversarially fine-tuned ResNet32.}
	\label{max perturbation table}
	\centering
	\begin{threeparttable}
		\begin{tabular}{r|ccccccc}
			\toprule
			\multicolumn{7}{c}{MNIST/CIFAR10}              \\
			\hline		
			& \multicolumn{6}{c}{Correctly classified test images (with mFI$\ge$0.01) by the  fine-tuned network of}\\
			Attacked	&mFI-PSO& FGSM&PGD& CW$_\infty$&JSMA&DeepFool\\
			by& (n=132/441) &  (n=169/570) &  (n=163/603) & (n=199/615) & (n=153/551) & (n=368/800) \\
			\cline{2-7}
			mFI-PSO & {\bf 44.06}/98.87 & 91.72/99.47 & 90.80/99.50 & 96.48/99.19 & 91.50/98.37 & 90.52/{\bf 97.75}\\\cline{2-7}
			FGSM	& 60.61/{\bf 29.02} & 81.82/29.47 & {\bf 52.76}/29.68 & 81.91/30.73 & 56.87/33.21 & 83.70/35.25\\\cline{2-7}
			PGD & {\bf 89.39}/32.43 & 100/30.88 & 98.16/{\bf 30.35} & 94.47/31.38 & 99.35/32.76 & 94.57/34.63\\\cline{2-7}
			CW$_\infty$  & {\bf 30.30}/{\bf 23.13} & 38.46/24.91 & 33.74/25.21 & 31.66/23.58 & 37.91/30.13 & 36.96/32.50\\\cline{2-7}
			\multirow{2}{*}{JSMA} & {\bf 0}/{\bf 24.49} & 0/43.68& 0/46.10 & 0/43.42 & 0/40.29 & 0/72.13\\
			&({\bf 98.48}/96.60) & (100/{\bf 96.32}) &({\bf 98.48}/98.34) & (100/96.91) & (100/96.91) & (100/97.63)  \\\cline{2-7}
			\multirow{2}{*}{DeepFool} & 15.91/5.90 & 10.06/{\bf 4.91} & 14.11/5.80 & 12.27/5.04 & 10.45/5.63 & {\bf 9.51}/8.13\\
			&({\bf 87.88}/{\bf 88.21}) & (89.35/91.40) & (89.57/89.88) & (87.94/91.54) & (91.53/88.93) & (91.30/90.13)
			\\
			\bottomrule
		\end{tabular}

		\begin{tablenotes}[flushleft]
			\item Note:  The preset bound of the $\ell_\infty$ norm of perturbations is
			$\epsilon=0.15$ for pixel values converted onto $[0,1]$. 
			JSMA and DeepFool do not have parameter $\epsilon$, and their success rates without 
			the $\ell_\infty$-norm bound are given in parentheses.
		\end{tablenotes}
	\end{threeparttable}
\end{table*}

We first investigate the success rates of the six attack methods.
We compare their attacks on the three subsets
of MNIST or CIFAR10
consisting of correctly classified images with mFI
$\ge\text{mFI}_{\text{img}}=0.2, 0.1, 0.01$, respectively.
We consider these vulnerable images rather than
all correctly classified images
due to the slow attack speed, e.g., of JSMA.
Table~\ref{attack succ table} reports the attack results.
The mFI measure indeed benefits
the success of our mFI-PSO attack.
Moreover, our mFI-PSO achieves the highest attack success rate on the subsets with threshold $\text{mFI}_{\text{img}}=0.2,0.1$
for MNIST
and those with $\text{mFI}_{\text{img}}=0.2,0.01$
for CIFAR10,
and also has a high rate
comparable to the best on the other two subsets.
In particular, our mFI-PSO wins  
with large margins ($\ge 24\%$) on 
the subset of CIFAR10 with $\text{mFI}_{\text{img}}=0.01$.
In contrast,
CW$_\infty$ has imbalanced performance on all the three MNIST subsets
with low success rates ($\approx 10\%$) on training data
but high rates ($> 97.5\%$) on test data.
Both JSMA and DeepFool have low success rates on
these selected images if with the constraint on the magnitude of perturbations.

We then compare the six methods in 
adversarially fine-tuning the newtork  
to build more robust classifiers. 
For each method,
the originally trained ResNet32 is additionally trained, 
with 30 epochs for MNIST and 80 epochs for CIFAR10,
on the combined set of original training data and its adversarial dataset from the subset with $\text{mFI}\ge 0.01$. 
For JSMA and DeepFool, we use their adversarial datasets including the images perturbed over the 0.15 $\ell_\infty$-norm bound. 
We randomly select 1/5 of the original training data and 1/5 of the adversarial dataset as the validation set for the fine tuning. 
Six fine-tuned ResNet32 models are obtained
for each of MNIST and CIFAR10.
Table~\ref{0.01acc adv table}
shows the
classification results of
each method's fine-tuned model
on the original test data and on its own adversarial data
from test images with $\text{mFI}\ge 0.01$.
All fine-tuned networks perform slightly better
than the originally trained network in accuracy on
the original test data, and have
a large improvement on corresponding adversarial datasets 
with accuracy over 82\% for MNIST and 
65\% (except DeepFool with 24.8\%) for CIFAR10.
For more fair comparison, 
the table lists the sample size of vulnerable test images with $\text{mFI}\ge 0.01$ for each fine-tuned network.
Our mFI-PSO dramatically reduces the sample size from 325 
and 1008 to 132 and 441 for MNIST and CIFAR10, respectively,
ranking the best among the six methods.

All adversarially fine-tuned networks
are further attacked by 
all the six methods.
Table~\ref{max perturbation table} shows the attack results
for each fine-tuned network on its correctly classified test images
that are vulnerable with $\text{mFI}\ge 0.01$.
The network fine-tuned by our mFI-PSO
generally exhibits the best defense performance
over the other five networks,
with most of
the lowest success rates of the six attack methods 
and with comparable results to the best rates on its unwon items. 
Moreover, for the other five fine-tuned networks,
our mFI-PSO attack still has high success rates over
90\% on both MNIST and CIFAR10, but the 
other five attacks have poor success rates 
less than 72.2\% (and below 46.2\% except JSMA) on CIFAR10, and 
CW$_\infty$, JSMA and DeepFool have rates below 38.5\% on MNIST.
For the network fine-tuned by mFI-PSO for CIFAR10,
the defense against mFI-PSO is not clearly seen on
its test subset of $\text{mFI}\ge0.01$,
but it is observed from the aforementioned dramatic reduced sample size (from 1008 to 441)
of the vulnerable test images that have $\text{mFI}\ge0.01$.
We also use mFI-PSO to attack
on the test subsets with $\text{mFI}\ge0.01$
from the original network but correctly classified
by the corresponding fine-tuned networks.
Enhanced defense against mFI-PSO
is observed
with an attack success rate 45.51\% for the mFI-PSO fine-tuned network,
in contrast to the worse rates between 60.83\% and 65.05\% for the other five networks.


\section{Conclusion}

This paper introduced a novel method called mFI-PSO for adversarial image generation for DNN classifiers by accounting for the user specified number of perturbed pixels, misclassification probability, and/or targeted incorrect class. We used a mFI measure
based on an ``intrinsic" perturbation manifold to efficiently detect the vulnerable images and pixels
to increase the attack success rate. We designed different misclassification loss functions to meet various user's specifications and obtained the optimal perturbation by the PSO algorithm. Experiments showed good performance of our approach in generating customized adversarial samples and associated adversarial fine-tuning for DNNs and its better performance in most studied cases over some widely-used~methods.

\bibliographystyle{IEEEtran}
\bibliography{paper}

\end{document}